\documentclass[conference]{IEEEtran}
\usepackage{cite}
\usepackage{amsmath,amssymb,amsfonts}

\usepackage{graphicx}
\usepackage{textcomp}
\usepackage{xcolor}
\usepackage{array}
\usepackage{tabularray}
\def\BibTeX{{\rm B\kern-.05em{\sc i\kern-.025em b}\kern-.08em
    T\kern-.1667em\lower.7ex\hbox{E}\kern-.125emX}}
\usepackage[most]{tcolorbox}
\usetikzlibrary{calc}
\usepackage[draft]{hyperref}
\usepackage{csquotes}
\usepackage{mathtools}
\usepackage{algpseudocode}
\newcommand{\semibold}[1]{{\small \textbf{#1}}}
\newcommand{\IMHERE}{{\semibold{IM HERE}}}
\newcommand{\EE}{{\textit{EE}}}
\newcommand{\EI}{{\textit{EI}}}
\newcommand{\Passive}{{\textit{Passive}}}
\newcommand{\Buildup}{{\textit{Buildup}}}
\newcommand{\Engaged}{{\textit{Engaged}}}
\newcommand{\Requested}{{\textit{Requested}}}

\DeclareMathOperator*{\argmax}{arg\,max}

\tcbset{
  mystyle/.style={enhanced, sharp corners, attach boxed title to top left, 
  colback=white, fonttitle=\small\bfseries,
  boxed title style={sharp corners, colback=gray!75!black,}}
}

\newtcolorbox[use counter=d]{definition}[2][] {mystyle, boxed title style={colback=blue!75!black},colframe=blue!75!black, label=#2, title=Definition \thetcbcounter,boxsep=0mm}
\newtcolorbox[use counter=i]{inference}[2][]  {mystyle,  boxed title style={colback=gray!75!black},colframe=gray!75!black, label=#2, title=Inference \thetcbcounter,boxsep=0mm, 
overlay unbroken and first={\node[anchor=south east] at (frame.south east) {\boxed{ #1 }};}, bottom=3ex,}
\newtcolorbox[use counter=a]{axiom}[2][]      {mystyle, boxed title style={colback=green!75!black},colframe=green!75!black,  label=#2, title=Axiom \thetcbcounter,boxsep=0mm}
\newtcolorbox[use counter=f]{foundation}[2][] {mystyle, boxed title style={colback=cyan!75!black},colframe=cyan!75!black,    label=#2, title=Foundation \thetcbcounter,boxsep=0mm}
\newcounter{d}
\newcounter{i}
\newcounter{a}
\newcounter{f}

\begin{document}

\title{IM HERE: Interaction Model for Human Effort based Robot Engagement\\}

\author{\IEEEauthorblockN{1\textsuperscript{st} Dominykas Strazdas}
\IEEEauthorblockA{\textit{Neuro-Information Technology} \\
\textit{Otto von Guericke University}\\
Magdeburg, Germany \\
dominykas.strazdas@ovgu.de}
\and
\IEEEauthorblockN{2\textsuperscript{nd} Magnus Jung}
\IEEEauthorblockA{\textit{Neuro-Information Technology} \\
\textit{Otto von Guericke University}\\
Magdeburg, Germany \\
magnus.jung@ovgu.de}
\and
\IEEEauthorblockN{3\textsuperscript{rd} Jan Marquenie}
\IEEEauthorblockA{\textit{Mobile Dialog Systems} \\
\textit{Otto von Guericke University}\\
Magdeburg, Germany \\
jan.marquenie@ovgu.de}
\and
\IEEEauthorblockN{4\textsuperscript{th} Ingo Siegert}
\IEEEauthorblockA{\textit{Mobile Dialog Systems}  \\
\textit{Otto von Guericke University}\\
Magdeburg, Germany \\
ayoub.al-hamadi@ovgu.de}
\and
\IEEEauthorblockN{5\textsuperscript{th} Ayoub Al-Hamadi}
\IEEEauthorblockA{\textit{Neuro-Information Technology} \\
\textit{Otto von Guericke University}\\
Magdeburg, Germany \\
ayoub.al-hamadi@ovgu.de}
}

\maketitle

\begin{abstract}


The effectiveness of human-robot interaction often hinges on the ability to cultivate engagement—a dynamic process of cognitive involvement that supports meaningful exchanges. Many existing definitions and models of engagement are either too vague or lack the ability to generalize across different contexts. We introduce \IMHERE, a novel framework that models engagement effectively in human-human, human-robot, and robot-robot interactions. By employing an effort-based description of bilateral relationships between entities, we provide an accurate breakdown of relationship patterns, simplifying them to \textit{focus} placement and four key \textit{states}. This framework captures mutual relationships, group behaviors, and actions conforming to social norms, translating them into specific directives for autonomous systems. By integrating both subjective perceptions and objective \textit{states}, the model precisely identifies and describes miscommunication. The primary objective of this paper is to automate the analysis, modeling, and description of social behavior, and to determine how autonomous systems can behave in accordance with social norms for full social integration while simultaneously pursuing their own social goals.

\end{abstract}



\section{Introduction}
Human interaction is hard to explain and even harder to model, yet we successfully interact with other humans, machines and objects every day. The field of Human-Robot Interaction (HRI) is broad,  encompassing many aspects of interaction, such as social robotics, human-robot collaboration, teaming, interfaces, autonomous systems, and many more. Throughout these aspects, a common challenge emerges: the need to recognize the current social states of the system and all the entities within it in order to determine appropriate actions and responses~\cite{strazdas2020robots}.

The concept of cognitive \textbf{engagement} facilitates smooth, natural communication between users and robots or agents, supporting meaningful, long-term interactions that extend beyond the initial novelty~\cite{salam2023automatic}.

Engagement, therefore, is not a static state but a continuous process of mutual adjustment, requiring entities to interpret, react to, and influence each other's behaviors. Understanding and modeling this process is essential for fostering interactions that are not only functional but also socially meaningful, enabling robots and agents to participate in human-like exchanges.

This dynamic requires not only accurate modeling but also the acknowledgment of reciprocity \cite{kelley1959social, duck1998human} as an additional concept in interaction. Reciprocity ensures that effort and engagement are mutual and perceived as balanced, thereby fostering social goals. For autonomous systems, this implies the need for deliberate revelation of their state to enable meaningful exchanges with human or machine counterparts.

In this paper, we propose a novel approach to engagement modeling by centering on a single sufficient indicator: \textbf{effort}. We argue that engagement fundamentally stems from bilateral \textit{focus}, and that effort can be used as an indicator. We develop a generalizable framework capable of predicting engagement for diverse entities, including humans, robots, and objects.

Our current framework, \IMHERE, is a theoretical concept to enhance existing engagement models by providing structure, precision, and general applicability without overriding them. Many current models lack a formal foundation or are scenario-bound, leading to ambiguity. It bridges subjective and objective views and revelation to understand communication and mitigate miscommunication. The framework integrates social norms in HRI, supports multi-party and group dynamics.
\IMHERE~enables the development of adaptive interaction strategies for goal-directed behavior, laying the foundation for real-world integration while accounting for noise and bias.
Additionally, we provide an open-source implementation of our model, along with a simulation environment designed to test different scenarios.

In the following sections, we explore existing engagement models and the various definitions of engagement used in different fields, distilling the foundational axioms and formulas needed for our framework. 

\section{Related Works}
Engagement is relevant to various scientific fields, mainly psychology \cite{gawronski2013dual} and computer science \cite{HCI_engagement_survey}. There are several definitions and models for engagement. In the context of our work, we focus on HRI.

\subsection{Definitions of Engagement}
There have been multiple approaches to model and predict engagement in HRI, summarized comprehensively by Oertel et al. \cite{oertel2020engagement} and Sorrentino et al. \cite{sorrentino2024definition}. Most of their reviewed literature uses the one of the following definitions for engagement:

\begin{itemize}

\item Sidner et al. \cite{sidner2003engagement} describe engagement as a dynamic process involving phases of initiation, maintenance, and conclusion. 

\item Poggi et al. \cite{poggi2007mind} view engagement as a value attributed to interaction partners by participants, which can be unidirectional.

\item Other concepts: The concept of “with-me-ness” \cite{lemaignan2016real} as the extent to which the human is “with” the robot. \cite{corrigan2013social}~refer to different types of engagement, such as task- or social-engagement and social-task-engagement e.g. collaboration. The term of group-engagement is introduced by \cite{salam2016fully}.
\end{itemize}

These definitions retain a certain level of ambiguity and vagueness, further examined by \cite{sorrentino2024definition,oertel2020engagement}.


\subsection{Modeling of Engagement}

Engagement is mainly viewed from the perspective of detection and prediction~\cite{ gao2020n, leite2015comparing, anzalone2015automated,anzalone2015evaluating, xu2013designing, bi2023human,kim2022joint,lu2024implementation,ben2019early, bohus2014managing, castellano2009detecting}, ignoring the reflection and revelation of engagement, that is necessary to develop interaction strategies.

These approaches utilize (multi-modal) indicators such as physiological signals \cite{gao2020n}, body posture \cite{ben2019early, leite2015comparing, anzalone2015automated, anzalone2015evaluating, xu2013designing, bi2023human}, facial expressions \cite{ben2019early, bohus2014managing, leite2015comparing, xu2013designing, castellano2009detecting}, gaze \cite{abdelrahman2022multimodal, hempel2023sentiment, ben2019early, anzalone2015automated, anzalone2015evaluating, castellano2009detecting, kompatsiari2021s, kompatsiari2019measuring,  leite2015comparing, nakano2010estimating, rich2010recognizing, xu2013designing}, head pose \cite{ben2019early, leite2015comparing, anzalone2015automated, anzalone2015evaluating, rich2010recognizing, xu2013designing, abdelrahman2022multimodal,hempel2023sentiment}, and speech patterns \cite{ben2019early, leite2015comparing, vaufreydaz2016starting, xu2013designing, bi2023human}. These models rely on specific indicators and lack a generalizable framework for diverse contexts, interaction types, and multi-person scenarios.

The use of such models may be sufficient in specific context, but for social robots, it is essential to not only detect the other parties state but to react and to reflect their own state through behavior and affective communication \cite{kelley1959social, duck1998human}. This can be subtle affirmations such as nodding \cite{lee2004nodding} or proactive in approaching another person \cite{hempel2023sentiment} or groups in F-formations \cite{joosse2014cultural}. 
Similarly, Obrien et al. \cite{obrien2008user} base their  engagement model around the user-experience, with different attributes (aesthetics, novelty, attention, positive affect, etc.) that lead to and prolong engagement and that should therefore be modulated by the system.

The goal of our research is to provide a fundamental, formal, generalizable concept for human-robot engagement, bridging the gap between definition and application scenario.
All aforementioned definitions of Engagement can be represented by our framework \IMHERE.

\section{\IMHERE: Effort based Engagement}

In the following, we present a brief overview of the proposed framework and the complexity of the relationship patterns it can model. Subsequent sections provide a rigorous formulation through definitions, axioms, and deductions.
\subsection{Framework Overview}
\label{sec:overwiew}
The framework is designed to represent the relational states — \Engaged, \Passive, \Requested, and \Buildup\ — between all entities from both objective and subjective perspectives. To start, the framework will be introduced using the relationship between two entities. For the remainder of the discussion, relationships between two entities will be described unless explicitly indicated otherwise.

\begin{figure}[h]
    \centering
    \includegraphics[width=.7\linewidth]{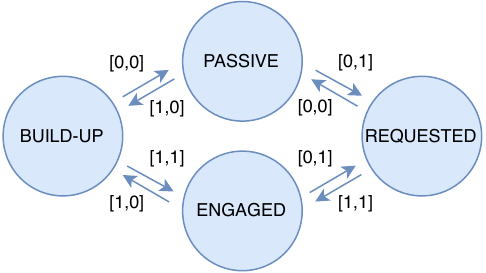}
    \caption{State machine illustrating relationship dynamics between two entities. Objects can only occupy \Passive\ and \Requested\ states, while engageable entities can transition through all four states. See Table~\ref{tab:relation_Entity-Entity} for mappings.}
    \label{fig:Statemachine}
\end{figure}
\vspace{-7mm}
\begin{table}[h]
    \centering
    \caption{Relationship between two Engageable Entities}
    
    \begin{tabular}{c|cc|c}
         Entity State&  Focus Entity&  Focus other Entity& Other Entity State\\
         \hline
         \Passive&  0&  0& \Passive\\
         \Requested&  0&  1& \Buildup\\
         \Buildup&  1&  0& \Requested\\
         \Engaged&  1&  1& \Engaged\\
    \end{tabular}
    
    \label{tab:relation_Entity-Entity}
\end{table}

At the heart of the proposed model, \IMHERE, is the concept of entity \textit{focus}. Based on the assumption that each entity can only maintain one \textit{focus} at a time, an \Engaged~\textit{state} is established between two entities when there is reciprocal \textit{focus}. The other states are reached, by different \textit{focus} combinations between two entities, as seen in  Figure \ref{fig:Statemachine}.
Each entity must estimate the \textit{focus} of others to accurately determine its own state. This is achieved by observing how others modulate their efforts, enabling optimization and better state estimation over time. By incorporating both subjective and objective perspectives, the model can capture various forms of miscommunication (divergence of subjective and objective views). In addition, the framework allows the identification of polite communication based on social norms by comparing the effort an entity uses to gain \textit{focus} with the effort that would be required according to social expectations. This can be used by autonomous systems to enhance its interactive capabilities in complex environments and effectively pursue its social goals.

The proposed framework consists of several stages: signal conversion, \textit{focus} estimation, state estimation, and goal evaluation, forming a closed-loop structure. These stages can be bypassed, modified, or added to. It functions as a modular toolkit capable of representing various social situations.

\subsection{Core Principles}
\IMHERE~uses a structured naming convention to clearly distinguish between definitions, foundational principles, axioms, and inferences, each of which contributes to the systematic development of the engagement framework.

\subsubsection{Definitions}
The framework begins by defining the essential components of engagement. Starting with the definition of those involved in interaction and communication: Entities. 

\begin{definition}{}{}
    An \textbf{Entity} is a \textit{self-contained physical object} capable of both sending and receiving signals, allowing it to participate in interactions.     
\end{definition}

Entities can therefore be people, animals, machines, or objects. The signals they actively or passively transmit to the environment, and thus to other entities, are referred to as effort.

\begin{definition}{d:effort}
    \textbf{Effort} represents the \textit{allocation of an entity’s limited} resources. 
\end{definition}
These limited resources are, for example, gaze, body or head orientation, movement (change of proximity), speech, etc. These resources are parametrized by alignment, magnitude and are susceptible to contrast.

The concept of attention is extensively utilized in literature as a descriptor of attentive engagement. Despite this, the term attention is only vaguely defined. The usefulness of the term Attention is challenged by Hommel and Colleagues in their paper "No one knows what attention is" \cite{hommel2019no} where they argue that the term Attention is not useful for the scientific community \cite{hommel2019no}. 
We instead employ the concept of focus, in line with other attention concepts \cite{lavie1995perceptual, posner1980orienting, treisman1980feature, treisman1964attenuation, broadbent1958perception}:

\begin{definition}{d:focus}
    \textbf{Focus} is \textit{a summary of an entity’s cognitive or perceptual resources} pointing towards one specific entity.
\end{definition}

It involves a deliberate or instinctive prioritization, where the entity allocates its sensory, mental, or operational capacity to process and respond to a particular object, action, or idea.
This act of selection creates a dynamic state in which the chosen target is emphasized in the entity's processing, enabling deeper interaction or understanding.

\subsubsection{Foundations}
The foundations are underlying axiomatic assumptions drawn from established theories to base the model upon. 
According to Watzlawick \cite{watzlawick1967pragmatics}  every entity is always communicating, and an entity "can not not communicate". 
As communication is part of engagement between entities, every entity is inherently emitting effort in different strengths to every other entity:
\\
\vspace*{-5mm} 
\begin{foundation}{f:communication}
    The \textbf{Communication Theorem} states that every action or inaction conveys a message ("You can not not communicate") \cite{watzlawick1967pragmatics}, implying that all entities inherently emit communication signals (Effort), consciously or unconsciously.
\end{foundation}

The second foundational concept is the individual interpretation of each effort by individual context, drawing from the Constructivist Theory of Communication \cite{von1984introduction}, Schramm’s Model of Communication \cite{schram1954process}, Helical Model of Communication \cite{dance1967helical}, Reception Theory\cite{barbatsis2004reception}, Transactional Model of Communication \cite{barnlund2017transactional} and Von Thun’s Four-Sides Model\cite{schulz1981miteinander}:

\begin{foundation}{f:subjectivity}
    \textbf{Subjectivity} acknowledges that communication signals are interpreted individually, as each entity perceives and evaluates signals within its own context. 
\end{foundation}

This subjectivity accounts for variability in responses and interpretations, making engagement a flexible, context-sensitive process, with high temporal dynamics. 
Together, these foundational principles highlight the omnipresence of effort and the variability in how it is perceived by each entity.

The basic concepts of \ref{d:effort} and \ref{f:communication} extend into:

\begin{inference}[\ref{d:effort},\ref{f:communication}]{i:EE}
        \textbf{Engagement Effort (EE)}, the \textit{measurable strength of a communication signal} sent from one entity.
\end{inference}
The outgoing engagement effort of an entity is changing on the way to their recipient by the environment. 
This signal is then interpreted by the receiver entity, as seen in \ref{f:subjectivity}:

\begin{inference}[\ref{d:effort},\ref{f:subjectivity}]{i:EI}
    \textbf{Effort Interpretation (EI)}  is the subjective interpretation of the effort from another entity sending out an engagement effort (\EE).
\end{inference}

The \textit{focus} of an entity points towards the entity with the \EE~that results in the strongest \EI.     

\begin{figure}[ht]
    \centering
    \includegraphics[width=0.5243299975\linewidth]{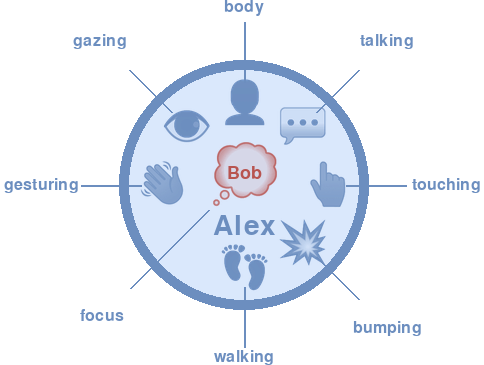}
    \caption{Entity \enquote{Alex} with 7 possible \EE~outputs, focusing on Entity \enquote{Bob}. 
    \label{fig:entity}}
\end{figure}


Figure~\ref{fig:entity} depicts an exemplary entity, as used in the framework simulation, with a name, position, orientation, field of view and size, as well as one \textit{focus} and different simulated \EE~output and \EI~capabilities. 
The outputs include:
\begin{description}
    \item[\textit{Walking}] ~~~~Producing footstep sounds.
    \item[\textit{Body}] ~~~~Representing physical presence.
    \item[\textit{Gaze}] ~~~~Indicating visual attention.
    \item[\textit{Touch}] ~~~~Representing intentional touching.
    \item[\textit{Gesture}] ~~~~Involving hand or body movements.
    \item[\textit{Talking}] ~~~~Conveying verbal effort.
    \item[\textit{Bumping}] ~~~~Indicating closeness (physical overlap).
\end{description}
The source code can be found \href{https://github.com/RoSA-Study/IM-HERE}{here}\footnote{\url{https://github.com/RoSA-Study/IM-HERE}}.

\subsubsection{Axioms}
Building on these definitions and foundations, the \IMHERE~model establishes three main principles governing all engagement dynamics. 

\begin{axiom}{a:focus}
    The \textbf{Focus Principle} asserts that an entity can direct its \textit{focus} to only one target at a time, specifically the one with the highest \EI.  
\end{axiom}
The single-focus constraint prevents divided attention and reinforces intentional, directed engagement (see \ref{d:focus}).

\begin{axiom}{a:engagement}
    The \textbf{Engagement Principle} specifies that mutual engagement only occurs when two entities have reciprocal \textit{focus}.
\end{axiom}
This axiom emphasizes that engagement is inherently mutual and occurs only between two parties. 

\begin{axiom}{a:revelation}
    The \textbf{Revelation Principle} requires an entities \textit{focus} to be observable, through \EE~modulation.
\end{axiom}
This means that an entity must consciously and purposefully direct effort toward the intended interaction partner for engagement to occur.
With these Definitions, Foundations and Axioms, our framework is fully defined and functional.

\subsection{Objects and Engageble Entities} 

To illustrate the functionality of \IMHERE, we now introduce an inference that differentiates between objects and engageable entities, clarifying their respective capabilities within the framework:

\begin{inference}[\ref{a:engagement},\ref{a:revelation}]{i:engageable} 
    An entity capable of differentiating between other entities and modulating \EE~is engageable.
\end{inference}


Objects, lacking the ability to interpret or modulate \EE, cannot actively focus on others. They are limited to the states \Passive~or \Requested. Engageable entities, by contrast, can navigate all states, including \Buildup~and \Engaged, enabling dynamic interactions with others. Building on this distinction, we delve into the stages and states of interaction that define how entities process and react to engagement efforts, forming the feedback loop essential for relationship modeling.

\subsection{Stages and States}
The \IMHERE~framework provides a flexible and adaptive approach for modeling relational states between entities. While we suggest a four-stage process as a foundation, this structure can be adapted or extended based on specific requirements. Designed to integrate seamlessly with existing models, it allows for modifications, such as combining stages or directly estimating \textit{focus} or engagement, without losing compatibility. This ensures that the framework can serve as both a standalone solution and a complement to established methodologies, enhancing the understanding of interaction dynamics.

\begin{figure}[ht]
    \centering
    \includegraphics[width=.7\linewidth]{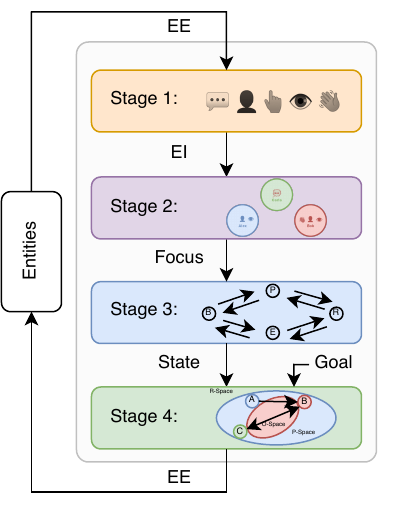}
    \caption{Stages of the framework within the feedback loop to optimize dynamic relationship estimation, social integration and the pursuit of social goals. 
    \label{fig:stages}}
\end{figure}

The proposed stages transition as follows: from incoming signals to \EI, from \EI~to \textit{focus}, from \textit{focus} to engagement \textit{states}, and finally from \textit{states} to goal-driven \EE~modulation. This \EE~is then perceived and processed by other entities, forming a dynamic feedback loop that enables continuous interaction and adaptation (see Figure\ref{fig:stages}).


\subsubsection{\textbf{Stage 1}}
In this stage, the incoming signal is interpreted by the receiving Entity and hence converted to EI. 
While there are multiple possibilities for \EI~calculation, we suggest a following approach that considers both influences by sender and the receiver and also the channel, through which the signal is being transmitted. The signal (\EE) has a magnitude $M$ and a contribution~$C$ which the sender can adjust. 

The alignment $A$ between the entities as well as the environment (contrast) $T$ have influence on this signal. Lastly, the personal preference $P$ of the receiver scales the signal. Every part of the signal is relative to the relation between sender and receiver. 
Mathematically, the \EI~calculation is simply a multiplication of those five variables:


\begin{gather}
     ~~~~~~ EI = T \cdot 
     \mathrlap{\overbrace{\phantom{M \cdot C  A ~)}}^{\text{sender}}}
      M \cdot C \cdot 
      \mathrlap{\underbrace{\phantom{A \cdot P}}_{\text{receiver}}} A \cdot P ~~~~~~~~~~~
    \makebox[\linewidth][l]{\footnotesize\begin{tabular}{ll} 
T & Contrast\\
M & Magnitude\\
C & Contribution\\
A & Alignment\\
P & Preference\\
    \end{tabular}}\notag
 \end{gather}

\begin{figure}[htbp] \centering \includegraphics[width=0.8\linewidth]{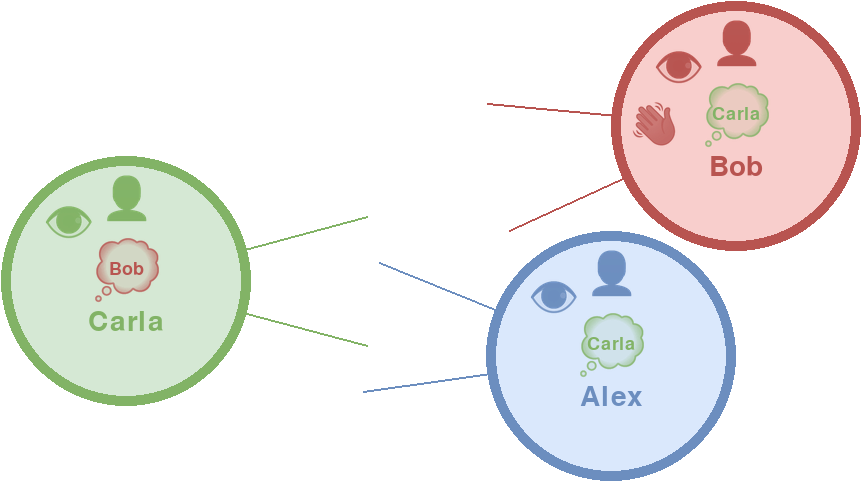} \caption{A, B, and C are interacting. A and C could have reciprocal \textit{focus}, but B is waving at C, sending a gesture \EE. C calculates the \EI s of A ({Body + Gaze}) and B ({Body + Gaze + Gesture}), and focuses on B, as his \EI~is highest.} \label{fig:EE-EI} \end{figure}

An example  of \EE~to \EI~calculation is illustrated in Figure~\ref{fig:EE-EI}. A (visual) gesture is sent from Bob (sender, \EE) to Carla (receiver, \EI). In this case, the alignment would depend on the relative position and orientation between the Entities, Magnitude would be the intensity of the gesture, contrast would be the relation between the intensity of the message and the background (noise, brightness, etc.), the contribution would contain the information of the message, and lastly the personal preference of the receiver. 

For Alex as receiver, Bob is standing outside his view, so the total signal strength is therefore rendered irrelevant, as the alignment is zero.  

\subsubsection{\textbf{Stage 2}}
An engageble entity can interpret multiple efforts on different channels (audio, visual etc.).
All \EI's coming from one sender can be cumulatively combined, thus resulting in total \EI. This is possible as the \EI~contains the subjective preferences of the receiver, which is used to scale the \EE~of each modality.
After estimating the $\EI_e$ of each entity ($e$) towards each every other entity including itself, the \textit{focus} of an entity can easily be calculated as the highest received \EI:
\begin{align}
    focus = \argmax_{e \in entities} \left( EI_{e} \right)
\end{align}


\subsubsection{\textbf{Stage 3}}

Combining the \textit{Focus-} \ref{a:focus} and \textit{Engagement Principle} \ref{a:engagement} with the \textit{Communication Theorem} \ref{f:communication} results in the engagement model and a corresponding state machine we introduced in section~\ref{sec:overwiew} for each entity.

The State of each Entity is determined by the \textit{focus}. If two entities ($A$, $B$) are \Engaged, they have reciprocal \textit{focus}:
\begin{align}
    Engaged_{A,B} \Leftrightarrow focus_A = B \land focus_B = A
\end{align}

When implementing this stage, the engagement status of an entity can be easily determined by checking whether the \textit{focus} of the \textit{focus} of an entity is the entity itself. Written as pseudocode:

\vspace{0.5em}{\noindent\fontsize{8.1}{9}\texttt{ entity.engaged = if (entity.focus.focus == entity)}}\vspace{0.5em}

Every relationship between two entities can be described by this state machine \ref{fig:Statemachine}. Only one state is possible at a time per relationship.   
To note: Objects can only be in the state of \Passive~or \Requested~as they have no \textit{focus}. 
In a broader environment with multiple entities, only the current \textit{focus} of an entity and if it is reciprocal needs to be traced (is there a relationship that is in state \Buildup~or in state \Engaged). For engageable entities, there is always exactly one relationship with state \Buildup or \Engaged.

\subsubsection{\textbf{Stage 4}}
The social goal of an entity always includes adopting a certain state in relation to another entity, therefore also manipulating the state of others. This can mean both engaging with a certain entity or avoiding being focused.

\begin{inference}[\ref{a:focus},\ref{i:EE}]{i:attraction}
    An entity can attempt to gain \textit{focus} by modulating its \EE.
\end{inference}
If \EE~is adjusted, the resulting \EI~may change, potentially becoming the highest for the target entity, thereby attracting their \textit{focus} (usually by increasing \EE).
Furthermore, \EE~is not only needed to become \Engaged, but also to stay \Engaged.

\subsection{Subjective vs Objective}
By tracking the \textit{focus} of all entities, the model can predict and simulate engagement dynamics in various scenarios objectively. This would be a meta-perspective, accurately projecting the states of engagement of multiple entities as the \textit{focus} of each entity is given.

Incorporating the second \textit{Subjectivity} Foundation~\ref{f:subjectivity} enables each entity to account for \EI~and infer others' \textit{focus} through their \EE, enhancing realism, adaptability, and nuanced engagement simulations.

The \textit{Revelation Principle}~\ref{a:revelation}  adds the reflection of the subjective perspective.
To achieve mutual engagement on a subjective level, when actively modulating an \EE, the other entity has to reflect \EE~to reveal its state and confirm the request:
\begin{inference}[\ref{f:subjectivity}\ref{a:revelation}]{i:effort_reflection} 
    Subjective engagement is established through mutual effort reflection.
\end{inference}
This could be a nod or a smile, a greeting, or just turning around to show the other person that you are listening. In reality there is no objective perspective, the objective perspective is an overlay of all subjective perspectives. 

\subsection{Quality of Engagement}
In any interaction, communication can be characterized as effective or polite within social norms:

\begin{definition}{}{}
    Communication is \textbf{effective} when the (objective) engagement is also perceived subjectively from both \Engaged~entities.
\end{definition}
Effective communication minimizes information loss.
Miscommunication arises when objective and subjective state estimations differ. By continuously analyzing \EE, estimating others' states, and comparing expected versus actual reactions, an entity can identify discrepancies between objective and subjective views and adjust its own estimation to reduce miscommunication.

Polite communication can be described using: 
\begin{definition}{}{}
    \textbf{Politeness} occurs when \EE~remains within social norms and adapts contextually.
\end{definition}
It estimates how much the \EE~must increase to gain or maintain the \textit{focus} of another entity. Underestimating this need can lead to a false perception of the situation, resulting in insufficient or overly reserved behavior. Conversely, overestimating it—such as shouting in a quiet setting—can come across as rude and uncomfortable. Moreover, while this may capture the target's \textit{focus}, it often draws unwanted attention from others in the environment.

\subsection{Engagement within Groups}
Group dynamics are highly complex; however, this model allows for the recognition of groups and their arrangement, as well as the pursuit of interaction goals based on this, such as joining the group or not disturbing it. Although only one entity at a time can be the focus of another, groups and group dynamics do not require a separate view. 
A group is characterized by a concatenation of \textit{focus}. 
For example, the \textit{focus} within a group is on the current speaker and the speaker's \textit{focus} alternates between entities within the group. The entire group is linked by their chain of \textit{focus}.

Group constellations can be easily recognized by following these \textit{focus} chains. Their spatial arrangement can then be described using F-formations. Thus, the inner O-space (reserved for activity), ring of P-space (place for interacting participants) and surrounding R-space (buffer to the outside world) can be determined directly from the \textit{focus} chaining and the spatial arrangement and allows third parties not to disturb the existing interaction by deliberately avoiding the O- and P-space or to integrate themselves by entering the P-space. 

\begin{figure}
    \centering
    \includegraphics[width=0.87\linewidth]{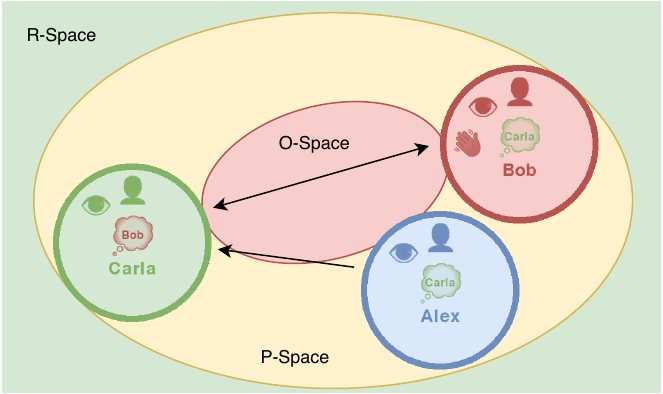}
    \label{fig:r_space}
    \caption{\textit{Focus} chaining and spatial arrangement in group dynamics, illustrating F-formations with O-space, P-space, and R-space.}
    
\end{figure}

An example of spatial arrangement in group dynamics of the previously discussed entities Alex, Bob and Carla, is shown in Figure~\ref{fig:r_space}.

\section{Implications for social robots} 
Social robots must be perceived as active participants capable of social engagement, rather than passive objects. To achieve this, robots need to act as engageable entities with clear goals and priorities in social contexts. 

\subsection{Prerequisites}

For a robot to estimate the focus of surrounding entities and define a virtual focus, it must receive signals (via sensors) and interpret them. To be perceived as engageable, it must emit an \EE~(via actuators) reflecting its \textit{virtual} \textit{focus}, akin to human behavior, such as a simple acknowledgment of presence. This enables \Buildup~and \Engaged~states, shifting the way it is perceived from an object (\Passive~or \Requested) to an active participant. A goal definition is essential for prioritizing \textit{focus}, whether on people seeking interaction or directing attention to a specific person.

\subsection{Strategy Pursuit}
The robot's goal setting requires it to modulate its \EE~in order to achieve the desired target state towards a specific entity, whether this is to communicate with those seeking assistance (\Passive~to \Requested~to \Engaged) or to interact with a particular entity (\Passive~to \Buildup~to \Engaged). This can also be applied for disengagement in the other direction, e.g. avoid interaction altogether (\Requested~to \Passive). By combining these basic strategies, the robot can interact within complex social structures.


\subsection{Miscommunication}

Miscommunication can occur if the robot misinterprets an \EE, such as mistaking a disengagement request or missing a \textit{focus} shift. This can often be mitigated by improving the robot’s \textit{focus} estimation and integrating modalities like gestures, tone of voice, or spatial positioning to enhance situational awareness. 

Errors can also stem from others misinterpreting the robot's state or intentions. If the robot's signals, such as gaze direction or posture, are unclear or inconsistent, others may misunderstand its \textit{focus} or willingness to engage. To prevent this, the robot must clearly and transparently reveal its internal state and intentions through deliberate and consistent cues.

\section{Discussion}

This framework is designed to be flexible and adaptable to other models. The formulas and mappings included in the stages are proposed as suggestions and can be specified to suit particular requirements or existing methodologies. Additionally, stages can be merged or simplified to align with established models while retaining the core principles and implications of the \IMHERE~framework.

It is not necessary to trace every step from raw signals to entity engagement (\EE) if certain states, such as \textit{focus} or engagement, can be directly estimated. This adaptability ensures that the framework provides a unifying structure capable of integrating with various approaches while offering novel insights and methodologies for engagement and interaction.

By adopting this flexible architecture, the \IMHERE~framework can serve both as a standalone model and as a complement to existing systems. It facilitates better understanding of social dynamics and supports the modulation of entity efforts within an interaction loop, enabling its application across diverse contexts and technologies.

\subsection{Specification and Recognition of Miscommunication}
A key contribution of the framework is its ability to differentiate between subjective and objective perspectives, allowing for the precise identification of miscommunication. By continuously estimating \EE, \EI, and \textit{focus}, the model detects when the subjective perception of one entity does not align with the objective state. 

\subsection{Integration of Existing Engagement Models}
The \IMHERE~framework harmonizes multiple perspectives from prior engagement definitions. By characterizing engagement as reciprocal \textit{focus} and incorporating subjective interpretations (\EI), it aligns with Sidner et al.'s \cite{sidner2003engagement} dynamic process model and Poggi et al.'s \cite{poggi2007mind} attribution-based perspective. 
Our framework addresses the vagueness and ambiguity of previous engagement models \cite{oertel2020engagement,sorrentino2024definition} by
incorporating cognitive engagement and modeling \textit{focus} shifts explicitly, bridging the gap between abstract definitions and their operationalization with the required high precision for human interaction.

\subsection{Flexibility in Application}
The framework's modular structure enables its application across various interaction scenarios, including:
\begin{itemize}     
    \item Task-oriented interactions: Robots can prioritize engagement with entities relevant to their operational goals.
    \item Social interactions: The model supports nuanced behavior modulation, enabling robots to exhibit politeness and adapt to social norms.
    \item Group dynamics: By analyzing \textit{focus} chains and F-forma\-tions, the framework facilitates group engagement and the integration of robots into multi-party interactions.
    \item Long-term engagement: Continuous optimization of \textit{focus} and \textit{effort} allows for sustained, meaningful interactions beyond initial novelty.
\end{itemize}

\subsection{Scalability and Complexity}
While the framework offers rich descriptive and predictive power, its complexity scales with the number of entities involved. Each entity has to estimate the \EE~of others, leading to the complexity of $O(N^N)$ in signal interpretation and \textit{focus} estimation. 
However, the binary nature of \textit{focus} (focused or not) simplifies state determination.
The complexity can be further reduced by focusing solely on relevant relationships.

\subsection{Limitations and Challenges}
Despite its strengths, the framework faces challenges:
\begin{itemize}
    \item Input data quality: Accurate estimation of \EE~and \textit{focus} depends on robust sensor data, including gaze direction, body posture, and environmental factors. Noise or ambiguity in these inputs may lead to incorrect state estimation.
    
    \item Real-world validation: The model's theoretical robustness must be tested in diverse real-world scenarios to evaluate its adaptability and effectiveness.
    
    \item Signal complexity: Multimodal signals (e.g., speech, gestures) require sophisticated processing pipelines to accurately model engagement dynamics.
\end{itemize}

\section{Conclusion}
In conclusion, several promising avenues for future work have emerged from this research. One significant direction involves developing optimization functions that can enhance \EE~modulation, tailored to specific contexts and interaction goals. 
Another vital step is the validation of the proposed framework, which includes implementing it in real-world HRI scenarios to refine the parameters and validate key assumptions. 
Finally, extending the applicability of this framework to diverse domains such as education, and healthcare offers an opportunity to explore its potential in settings where sophisticated engagement is particularly crucial.

The \IMHERE~framework introduces a novel perspective on engagement in HRI, emphasizing reciprocity, \textit{focus} dynamics, \textit{effort}, and revelation as the fundamental components. This approach extends beyond existing models by offering a comprehensive and generalizable structure for analyzing and modeling engagement. 
Its modularity, generalizability, and ability to address miscommunication make it well-suited for autonomous systems striving to integrate seamlessly into human social environments.
This foundational framework presents yet another step towards natural HRI.


\section*{Acknowledgments} 
This work is funded and supported by the Investment Bank of Saxony-Anhalt (IB-Sachsen-Anhalt) (AI Co-Working Labs under grant No. ZS/2024/01/183362), by the German Research Foundation (DFG) (SEMIAC under grant number No. 502483052), by the Federal Ministry of Education and Research of Germany (BMBF) (AutoKoWAT-3DMAt under grant No. 13N16336) and by the European Regional Development Fund (ERDF) (ENABLING under grant No. ZS/2023/12/182056).

\bibliographystyle{IEEEtran}
\bibliography{references}

\end{document}